\documentclass[letterpaper]{article} 
\usepackage{aaai23}  
\usepackage{times}  
\usepackage{amsmath}
\usepackage{amsfonts}
\usepackage{amsthm}
\usepackage{diagbox}
\usepackage{multirow}
\usepackage{helvet}  
\usepackage{courier}  
\usepackage[hyphens]{url}  
\usepackage{graphicx} 
\usepackage{natbib}  
\usepackage{caption} 
\usepackage{algorithm}
\usepackage{algorithmic}
\usepackage{wasysym}
\usepackage[misc]{ifsym}
\usepackage{newfloat}
\usepackage{listings}
\DeclareCaptionStyle{ruled}{labelfont=normalfont,labelsep=colon,strut=off} 
\floatstyle{ruled}
\newfloat{listing}{tb}{lst}{}
\floatname{listing}{Listing}
\title{Measuring the Privacy Leakage via Graph Reconstruction Attacks on Simplicial Neural Networks (Student Abstract)}
\author{
   Huixin Zhan\textsuperscript{\rm 1\textdagger},
   Kun Zhang\textsuperscript{\rm 2\textdaggerdbl},
   Keyi Lu\textsuperscript{\rm 3\textsection},
   and Victor S. Sheng\textsuperscript{\rm 1\textasteriskcentered }}
\affiliations{
    \textsuperscript{\rm 1} Department of Computer Science, Texas Tech University, Box 43104, Lubbock, TX 79409-3104, USA\\
\textsuperscript{\rm 2} Department of Computer Science, Xavier University of Louisiana, New Orleans, LA 70125, USA \\
\textsuperscript{\rm 3} Department of Computer Science and Engineering, The Ohio State University, Columbus, OH 43210, USA \\
    \{\textsuperscript{\textdagger}Huixin.Zhan, \textsuperscript{\textasteriskcentered}Victor.Sheng\}@ttu.edu, \textsuperscript{\textdaggerdbl}kzhang@xula.edu, \textsuperscript{\textsection}lu.2285@osu.edu, 
}

\usepackage{bibentry}

\begin{document}

\maketitle

\begin{abstract}
In this paper, we measure the privacy leakage via studying whether graph representations can be inverted to recover the graph used to generate them via graph reconstruction attack (GRA). We propose a GRA that recovers a graph's adjacency matrix from the representations via a graph decoder that minimizes the reconstruction loss between the partial graph and the reconstructed graph. We study three types of representations that are trained on the graph, i.e., representations output from graph convolutional network (GCN), graph attention network (GAT), and our proposed simplicial neural network (SNN) via a higher-order combinatorial Laplacian. Unlike the first two types of representations that only encode pairwise relationships, the third type of representation, i.e., SNN outputs, encodes higher-order interactions (e.g., homological features) between nodes. We find that the SNN outputs reveal the lowest privacy-preserving ability to defend the GRA, followed by those of GATs and GCNs, which indicates the importance of building more private representations with higher-order node information that could defend the potential threats, such as GRAs.
\end{abstract}

\section{Introduction}
Most real-world graphs associated with people or human-related activities are often sensitive and might contain confidential information. In this paper, we focus on the threat of edge privacy, e.g., in an online social network, a user's friend list could potentially be private to the user. The server could aggregate node representations with their neighbors to learn better user representations to improve its services. Thus, if there is an edge between two nodes, their output representations obtained from the graph neural network (GNN) should be closer. Therefore, a potential adversary could possibly recover the sensitive edge information (e.g., friend lists) via a machine learning classifier that computes distance differences in graph representations.

In this paper, we study three representation methods (two low-order and one higher-order) that perform different aggregations to capture important graph structure properties, i.e., representations output from GCN, GAT, and our proposed SNN. Unlike the first two types of representations that only encode pairwise relationships, the proposed SNN utilize a higher-order combinatorial Laplacian to learn a graph convolution that encodes the homological features of simplicial complexes, which are higher-dimensional analogs of graphs~\citep{horak2013spectra}. We will then measure if SNN outputs are more vulnerable to potential attacks compared to regular GCN or GAT outputs.


We use GRAs as the potential attacks to measure privacy leakage. GRAs study whether representations can be inverted to recover the graph used to generate them. We propose a novel GRA for accurate representation inversion, i.e., from the representation outputs of a graph $\mathcal{G}$, we can find a graph $\hat{\mathcal{G}}$ with a very similar adjacency matrix. When a user’s device computes a representation via GNN and sends it to a server for node classification, we assume the adversary could access the representation outputs with non-confidential edges during the data uploading process. We propose a graph decoder that reconstructs the graph by minimizing the reconstruction loss between the partial graph (with non-confidential edges) and the reconstructed graph. In our experiments, we will show that the SNN outputs reveal the lowest privacy-preserving ability to defend the GRAs. This calls for future research towards building more private representations with higher-order node information that could defend the potential attacks, such as GRAs. 

\section{Our Proposed Methods}
\begin{table*}[!ht]
\centering

\begin{tabular}{l| c c c c c } 
 \hline
\backslashbox[40mm]{Attacks}{Dataset}&  Citeseer&   Cora& Pubmed & Computers & Photos \\ [0.5ex]
 \hline
 Attack-6~\citep{he2021stealing}  & $0.9795$  & $0.9638$  & $0.9702$  & $0.9800$  & $0.9691$\\
 GRA\_GCN outputs (Ours)& $0.8832^\downarrow$  & $0.8900^\downarrow$   & $0.8937^\downarrow$& $0.9041^\downarrow$  & $0.9157^\downarrow$\\
GRA\_GAT outputs (Ours)& $0.9224^\downarrow$  & $0.9640$  & $0.9704$  & $0.9800^{=}$  & $0.9703$\\ 
\textbf{GRA\_SNN outputs (Ours)} & $\mathbf{0.9833}$  & $\mathbf{0.9713}$   & $\mathbf{0.9826}$  & $\mathbf{0.9802}$  & $\mathbf{0.9832}$\\ 
 \hline
\end{tabular}
\caption{AUC Score on Three Types of Representations.}
\label{table:recall}
\end{table*}
\paragraph{SNN Outputs}
In this paper, we will refer abstract simplicial complex to simplicial complexes. An abstract simplicial complex is a collection $K$ of subsets of a finite set $S$ that satisfies two axioms: (1) The singleton set $\{v\}$ lies in $K$ for each $v$ in $S$. (2) Whenever some $\sigma \subset  S$ lies in $K$, every subset of $\sigma$ must also lie in $K$. The constituent subsets $\sigma \subset S$ which lie in $K$ are called simplices. Next, we will introduce the computation of $d$-dimensional incidence matrices. Fixing a collection $K$ and letting $K_d$ indicate the set of all $d$-simplices in $K$, the $d$-dimensional incidence matrices operators can be represented as $ \partial_d: \mathbb{R}^{K_d} \rightarrow \mathbb{R}^{K_{d-1}}$. To build these incidence matrices operators, one first orders the nodes in $K_0$ so that each $d$-simplex $\sigma \in K$ can be uniquely expressed as a list $\sigma = [v_0,..., v_d]$ of nodes in increasing order. The desired matrix $\partial_d$ is completely prescribed by the following action on each $\sigma$: 
$\partial_d (\sigma) = \sum_{i=0}^d (-1)^i \cdot \sigma_{-i},$
where $\sigma_{-i} :=  [v_0,...,\hat{v}_i,..., v_d]$ and $\hat{v}_i$ indicates that $v_i$ is omitted. These operators form a sequence of vector spaces and linear maps:
\begin{equation}
\cdots \overset{\partial_{d+1}}{\rightarrow} \mathbb{R}^{K_d}  \overset{\partial_{d}}{\rightarrow} \mathbb{R}^{K_{d-1}}  \overset{\partial_{d-1}}{\rightarrow}\cdots. 
\end{equation}
In order to model the higher-order interactions between nodes, the graph Laplacian was generalized to simplicial complexes by~\citet{horak2013spectra}. The higher-order combinatorial Laplacian can be formulated as:
$
\mathcal{L}_d := \partial_{d+1} \partial_{d+1}^T + \partial_{d}^T \partial_{d}.$ Therefore, leveraging this $\mathcal{L}_d,$ we could obtain the final SNN outputs via the graph convolution as follows:
$
H^{(2)} = \sigma(\tilde{Q}^{-\frac{1}{2}}\tilde{\mathcal{L}}_d\tilde{Q}^{-\frac{1}{2}} \sigma(\tilde{Q}^{-\frac{1}{2}}\tilde{\mathcal{L}}_d\tilde{Q}^{-\frac{1}{2}}H^{(0)}W^{(0)})W^{(1)}),\notag
$
where $\tilde{\mathcal{L}}_d = \mathcal{L}_d+\mathbb{I}$, $\mathbb{I}$ is the identity matrix, $\tilde{Q}_{ii} = \sum_j\tilde{\mathcal{I}}_{d_{ij}}$ and $W^{(k)}$ is a layer-specific trainable weight matrix. $\sigma(\cdot)$ denotes an activation function. $H^{(k)}$ is the matrix of activations in the $k$-th layer and the initial node representations are $H^{(0)}$. 

\paragraph{Graph Reconstruction Attack}
A graph $\mathcal{G} = (V, E)$ is represented by the set of nodes $V = \{v_i\}_{i = 1}^{|V|}$ and edges $E = \{e_{ij}\}_{i,j = 1}^{|E|}$. To measure the privacy leakage via GRA, we will then propose a decoder only approach. Specifically, we utilize the outputs $H^{(k)}$ and a partial adjacency matrix $\mathcal{A}^{*}$ as the prior knowledge, the decoder $f_{dec}$ reconstructs the adjacency matrix $A_{rec} = f_{dec}(H^{(k)})$ via 
$
A_{rec} = \sigma((\tilde{\mathcal{A}}^{*} H^{(k)} W_a) (\tilde{\mathcal{A}}^{*} H^{(k)} W_a)^\mathbf{T}),
$
where $W_a$ is trained using back-propagation to minimize reconstruction loss between the adjacency matrix of thee partial graph $\mathcal{A}^{*}$ and the reconstructed partial graph $\mathcal{A}_{rec}^{*}$ using 
$
\mathcal{L}^{(att)} = ||\mathcal{A}^{*}-\mathcal{A}_{rec}^{*}||^2_2.$

\section{Experiments}

\paragraph{Datasets}
In our experiments, we used five well-known real-world datasets:  CiteSeer and CORA from~\citet{sen2008collective}, PubMed~\citep{namata2012query}, as well as Amazon Computers and Amazon Photo from~\citet{shchur2018pitfalls}. 
\paragraph{Experimental Set-Up}
In order to generate GCN and GAT outputs, we follow the original graph convolution in GCNs~\citep{welling2016semi} and concatenation in GATs~\citep{velivckovic2017graph}. We only consider $1$-simplices, i.e., $d=1.$ 
\paragraph{Performance for Three Types of Representations}

Table~\ref{table:recall} shows the GRA performance in terms of AUC of three types of representations for recovering the adjacency matrix $\mathcal{A}$ of $\mathcal{G}$. We compare our results with Attack-6~\citep{he2021stealing} because it also requires node representations and the partial adjacency matrix $\mathcal{A}^{*}$ as prior knowledge. Our proposed GRA with GCN outputs (GRA\_GCN outputs) obtains inferior performances compared to Attack-6 (with $\downarrow$). However, GRA\_GAT outputs achieve higher performances compared to Attack-6 in most of the cases and GRA\_SNN outputs achieve the highest performances for all datasets (in bold), e.g., it achieves $98.33\%$ AUC on the Citeseer dataset. This indicates the SNN outputs reveal the lowest privacy-preserving ability to
defend the GRAs, followed by GAT and GCN outputs.

\section{Conclusion}
In this paper, we measure the privacy leakage via studying whether representations can be inverted to recover the graph used to generate
them using GRA. Our proposed GRA recovers a graph’s adjacency matrix from the representations via a graph decoder that minimizes the reconstruction loss. We studied three types of representations that are trained on the graph, i.e., GCN outputs, GAT outputs, and SNN outputs using a higher-order combinatorial Laplacian. We found SNN outputs reveal the lowest privacy preserving ability. This indicates
the importance of building more private representations with higher-order node information that could defend the potential threats. 

\bibliography{aaai23}

\begin{thebibliography}{7}
\providecommand{\natexlab}[1]{#1}

\bibitem[{He et~al.(2021)He, Jia, Backes, Gong, and Zhang}]{he2021stealing}
He, X.; Jia, J.; Backes, M.; Gong, N.~Z.; and Zhang, Y. 2021.
\newblock Stealing links from graph neural networks.
\newblock In \emph{Proceedings of the 30th USENIX Security Symposium},
  2669--2686.

\bibitem[{Horak and Jost(2013)}]{horak2013spectra}
Horak, D.; and Jost, J. 2013.
\newblock Spectra of combinatorial Laplace operators on simplicial complexes.
\newblock \emph{Advances in Mathematics}, 244: 303--336.

\bibitem[{Namata et~al.(2012)Namata, London, Getoor, Huang, and
  EDU}]{namata2012query}
Namata, G.; London, B.; Getoor, L.; Huang, B.; and EDU, U. 2012.
\newblock Query-driven active surveying for collective classification.
\newblock In \emph{10th International Workshop on Mining and Learning with
  Graphs}, volume~8, 1.

\bibitem[{Sen et~al.(2008)Sen, Namata, Bilgic, Getoor, Galligher, and
  Eliassi-Rad}]{sen2008collective}
Sen, P.; Namata, G.; Bilgic, M.; Getoor, L.; Galligher, B.; and Eliassi-Rad, T.
  2008.
\newblock Collective classification in network data.
\newblock \emph{AI Magazine}, 29(3): 93--93.

\bibitem[{Shchur et~al.(2018)Shchur, Mumme, Bojchevski, and
  G{\"u}nnemann}]{shchur2018pitfalls}
Shchur, O.; Mumme, M.; Bojchevski, A.; and G{\"u}nnemann, S. 2018.
\newblock Pitfalls of graph neural network evaluation.
\newblock \emph{arXiv preprint arXiv:1811.05868}.

\bibitem[{Veli{\v{c}}kovi{\'c} et~al.(2017)Veli{\v{c}}kovi{\'c}, Cucurull,
  Casanova, Romero, Lio, and Bengio}]{velivckovic2017graph}
Veli{\v{c}}kovi{\'c}, P.; Cucurull, G.; Casanova, A.; Romero, A.; Lio, P.; and
  Bengio, Y. 2017.
\newblock Graph attention networks.
\newblock \emph{arXiv preprint arXiv:1710.10903}.

\bibitem[{Welling and Kipf(2016)}]{welling2016semi}
Welling, M.; and Kipf, T.~N. 2016.
\newblock Semi-supervised classification with graph convolutional networks.
\newblock In \emph{Proceedings of the 5th International Conference on Learning
  Representations}.

\end{thebibliography}

\end{document}